\documentclass[journal]{IEEEtran}
\usepackage{graphicx}
\usepackage{comment}
\usepackage[cmex10]{amsmath}
\usepackage{amssymb}
\usepackage{color}
\usepackage{subfigure}
\usepackage{subcaption}
\usepackage{array}

\usepackage{algorithm}
\usepackage{algorithmicx}%
\usepackage{algpseudocode}%

\begin{document}%

\title{Accelerated Multi-objective Task Learning using Modified $Q$-learning Algorithm}

\author{Varun Prakash Rajamohan, Senthil Kumar Jagatheesaperumal
\thanks{V.P. Rajamohan, S.K. Jagatheesaperumal are with Department of Electronics and Communication Engineering, Mepco Schlenk Engineering College, Sivakasi-626005, Tamil Nadu, India \\
E-mail: varunprakash.r@mepcoeng.ac.in, senthilkumarj@mepcoeng.ac.in}}
%

%
%
%
%
%
%
%
\maketitle

\begin{abstract}
Robots find extensive applications in industry. In recent years, the influence of robots has also increased rapidly in domestic scenarios. The Q-learning algorithm aims to maximise the reward for reaching the goal. This paper proposes a modified version of the $Q$-learning algorithm, known as $Q$-learning with scaled distance metric ($Q-SD$). This algorithm enhances task learning and makes task completion more meaningful. A robotic manipulator (agent) applies the $Q-SD$ algorithm to the task of table cleaning. Using $Q-SD$, the agent acquires the sequence of steps necessary to accomplish the task while minimising the manipulator's movement distance.  We partition the table into grids of different dimensions. The first has a grid count of $3 times 3, and the second has a grid count of $4 times 4. Using the $Q-SD$ algorithm, the maximum success obtained in these two environments was 86\% and 59\% respectively. Moreover, Compared to the conventional $Q$-learning algorithm, the drop in average distance moved by the agent in these two environments using the $Q-SD$ algorithm was 8.61\% and 6.7\% respectively.
\end{abstract}

\begin{IEEEkeywords}Q-Learning; Cleaning scenario; Task planning; Area Coverage; Multi-objective
\end{IEEEkeywords}





\maketitle

\section{Introduction}\label{Intro}
Robots are extensively used for object manipulation in various applications, ranging from industry to domestic settings. As population density increases, there is a growing demand for task planning and cleaning robots \cite{cleaningRobotIntro_1}. These robots play a crucial role in both industrial and domestic environments. Regular cleaning is essential for maintaining the living standards and functionality of these buildings \cite{cleaningRobotIntro_2}. Therefore, the automation of cleaning processes in building infrastructures has become increasingly necessary. To achieve efficient automation, robotic agents need to be equipped with intelligent algorithms to perform tasks effectively. Robots installed in buildings to autonomously complete area coverage tasks like cleaning, building inspection and so on \cite{lakshmanan2020complete}. Infants acquire object manipulation skills at an early stage through interaction. Similarly, Reinforcement Learning (RL) solves problems by making repeated attempts with appropriate rewards. A wide variety of robotic applications also make use of RL\cite{TanCheeReview}. In each instance of interaction, RL performs an action that causes a change in the environment's state. The agent is rewarded accordingly based on the state change. Through repeated actions, the agent maximizes the cumulative reward collected \cite{QL_similarSutton}. $Q$-learning algorithm is a table-based method with better convergence. In this work, the $Q$-learning algorithm is modified to learn tasks with reduced distance moved. This algorithm is applied to the task of cleaning the table partitioned as grids. 

Cristian et al. \cite{MillnArias2021ARA} proposed three different approaches to object handling by the manipulator. They are soft actor-critic-based Interactive RL, Robust RL, and Interactive Robust RL. These different techniques were implemented in a simulation environment of manipulator classifying objects. In this work, advice is given related to the task and dynamics of the environment. As a result, performance is better in IRL compared to the classic RL algorithm. The training episodes in IRL are reduced compared to classic RL. RRL has taken more episodes for training but attains good performance with external disturbances. IRRL is better than RRL in terms of training. 

Cruz et al. \cite{IRLbase7458195} proposed a learning algorithm for object handling in simulated domestic scenarios. In this, the agent manipulates objects like obstacles and sponges to clean the table. This problem was solved using three different algorithms namely classic RL, RL with affordance, and Interactive RL. In classic RL success rate of 35\% was obtained with training of thousand episodes. In the second and third approaches, the number of episodes for training was reduced to a hundred. In IRL even receiving small advice around 10\% enables the robot to finish the cleaning task faster. This work has better convergence compared to \cite{ImprRLAffrdrefBase6982975}. However, in this work, the focus is on efficient task learning with no adequate attention towards the utility value. Moon et al.~\cite{moon2022path} applied proximal policy optimization(PPO) combined with RL for cleaning using a mobile robot. They obtained a better performance in comparison to conventional methods. In \cite{zhang2023predator} $ Q$-learning-based Coverage Path Planning (CPP) is proposed. Compared to the conventional $Q$ learning algorithm uses a predator-prey method to reward allocation. Thereby it avoids the problem of local optima. Additionally, it decreases the repetition ratio and converts numbers in CPP. This can be improved even more by applying it to an environment involving dynamic obstacles. Knowledge about handling objects in one task is transferred to other tasks in \cite{KnowledgeTfrBoloka}. Here probability-based policy reuse is combined with with Q-learning. Learning time for new tasks is reduced due to knowledge transfer. This algorithm is applied to basic object-handling mechanisms by the robot and can be improved towards in-task learning processes. Further, this work can be improved to handle high sensory input.

Handling of objects in cluttered environments with multiple objects was discussed in \cite{Cheong2021ObstacleRF} and \cite{InteractivePerceptionWarehouse8970574}. Cheong et al.\cite{Cheong2021ObstacleRF} proposed the manipulation of obstacles to find a collision-free path to the target object. Objects are arranged in a grid-based environment. Deep Q-network is the learning algorithm used to learn which object to pick and where to place. Two different deep learning architectures Single DQN and sequentially separated DQN were proposed. This work attains a reduction in execution time and number of obstacles rearranged by up to 35\%. Sequentially separated DQN has better performance as the number of obstacles increases. This can be further improved by considering the agent's effort for rearranging and also taking into consideration uncertainty in the environment. This can be enhanced by taking into account the agent's labour for reorganizing and also addressing the ambiguity in the environment. Similarly, in \cite{InteractivePerceptionWarehouse8970574} proposed an interactive perception method for object grasping in a cluttered environment. Initially affordance map of the environment was obtained from an RGB-D image. This affordance map tells about the confidence of each pixel for grasping. If the affordance map is inappropriate, then push action is done until a suitable affordance map is obtained. The exploration strategy is based on deep-RL. This work has a better suction success rate and scene success rate. A similar approach is proposed in \cite{DeepRLRoboticPushingClutteredDeng} but without tactile sensing. Compared to the work in \cite{Cheong2021ObstacleRF}, environmental uncertainty has been handled, but agent effort for the agent's labor has not been quantified. 

RL in application towards coverage of area is applied in below listed papers below. The problem of area coverage is addressed in \cite{xiao2020distributed}. RL and $\gamma$-information map are used for the distributed dynamic area coverage algorithm. They proposed a $Q$-traversal algorithm based on the  $\gamma$ and distributed cooperative $Q$-learning algorithm. However, its efficiency is constrained in case of communication distance between the agents is very large. Further, the focus is focused towards coverage and utility values like energy spent in coverage or distance travelled for coverage are not considered. Anirudh Krishna et al., \cite{lakshmanan2020complete} proposed a coverage planning algorithm that was implemented using deep black RL for reconfigurable robotic platforms. This work generates paths with reduced cost and time. In the work \cite{thakar2022area} coverage area of the surface from the spray of the nozzle was addressed. It is applied to the task of disinfection. It uses an area coverage planning algorithm to compute the path of the nozzle to cover the area. Though it is better in terms of better coverage and disinfectant wastage the performance drops for large depth areas. In the work \cite{piardi2019coverage} the area is partitioned into grids with known obstacle positions. And $Q$-learning is applied to the problem of CPP.  Better coverage is obtained by using Deep RL in the work \cite{jonnarth2023end}. However, in this work map representation of the environment is used which may increase the computation complexity. Coverage of an uneven surface using automatic segmentation is given in \cite{Schneyer}, but additional utility values have not been employed. The coverage of 3-D surface is proposed in the work \cite{mcgovern2023general}.

Compared to other works the objective of this work is task learning with added utility value.  To this end, we propose a modified $Q$-learning algorithm namely $Q$- learning with scaled distance metric ($Q-SD$).
By using the $Q-SD$ algorithm the agent not only learns the task of cleaning a table with multiple objects but also accomplishes the task with reduced distance moved by the agent. The main contributions of the work are, 
\begin{itemize}
    \item Q-learning algorithm is modified to develop a $Q-SD$ algorithm. 
    \item The agent's learning rate is improved by incorporating appropriate weight to the scale factor of the distance metric.
    \item Impact analysis of scaled distance metric on the task learning of the agent.
    \end{itemize}
The rest of this work is organized as a description of the agent's learning environment and reward allotment is described in section \ref{framework}. The design and algorithmic description of the $Q-SD$ algorithm is given in section \ref{rlBasic}. The obtained result and its analysis are given in section \ref{simltnResults}. Finally, the conclusion and challenges are described in section \ref{conclusion}.

\section{Framework of Agent's Learning Environment} \label{framework}

In this work, the problem of table cleaning with objects at the centre is being considered. The objective is to clean the table area without objects. This work proposes a novel $Q-SD$ algorithm to learn the task of cleaning the table with the minimum distance moved by the agent. The entire table area is discretized into $G$ number of grids. Each grid is taken to be of identical size and the object solely occupies a single grid. The robotic arm is taken as capable of reaching any grid on the table. 

\begin{figure}[t]%
\centering
\includegraphics[width=0.4\linewidth]{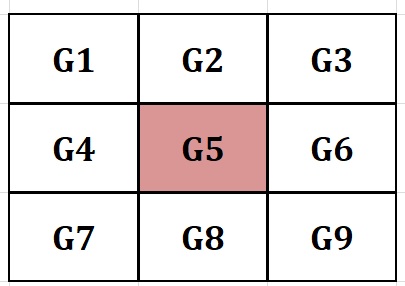}
\caption{$3\times3$ Grid with objects placed at center ($G5$). }\label{fig:Grid3x3}
\end{figure}

\begin{figure}[t]
\centering
\includegraphics[width=0.7\linewidth]{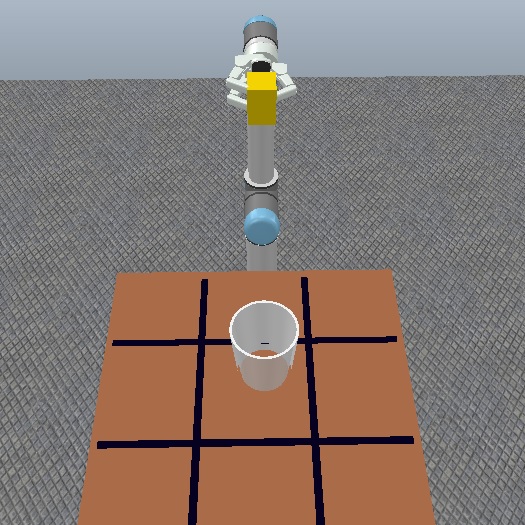}
\caption{Simulation environment illustrating a scenario with a grid of dimensions $3\times3$ done using CoppeliaSim  \cite{rohmer2013v}}\label{fig:simGrid3x3}
\end{figure}

\subsection{Affordance Environment Definition - Table with Grid}\label{gridScenario}
\begin{figure}[ht]%
\centering
\includegraphics[width=0.65\linewidth]{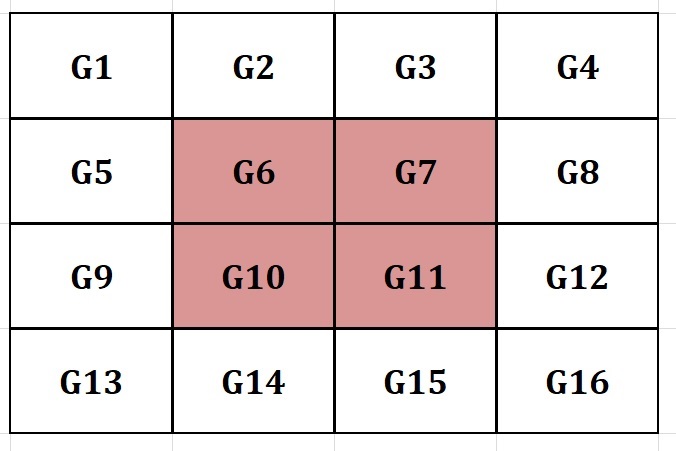}
\caption{$4\times4$ Grid with objects placed at center ($G6, G7, G10, G11$)}\label{fig:Grid4x4}
\end{figure}

\begin{figure}[h]%
\centering
\includegraphics[width=0.65\linewidth]{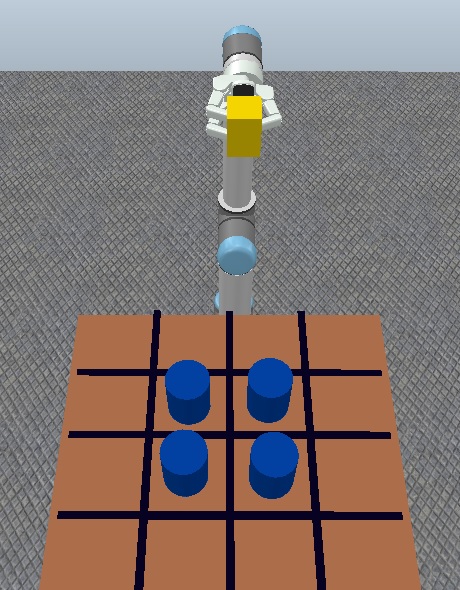}
\caption{Simulation environment illustrating a scenario with a grid of dimensions $4\times4$ done using CoppeliaSim  \cite{rohmer2013v}}\label{fig:simGrid4x4}
\end{figure}

 Two different grid partitions were considered one is of size $3\times3$ and the other is of size $4\times4$. In the grid of size, $3\times3$ nine grids range from $G1$ to $G9$. The object is placed in $G5$ as depicted in figure \ref{fig:Grid3x3}. The size of the object is taken to be within one grid size. The agent is trained to clean the grid in the table that contains no object. The terminal state is all the grids without objects has been cleaned. The simulation scenario of the $3\times3$ grid is given in the figure. \ref{fig:simGrid3x3}. The state table for $3\times3$ is described in table \ref{tab:state3x3}. For this environment design it consists of $256$ states. Similarly, for the grid of size $4\times4$ sixteen grids range from $G1$ to $G16$. Among those grids, $G6, G7, G10$ and $G11$ are occupied with objects as described in figure \ref{fig:Grid4x4}. Here again, objects are taken to be occupying the centre part of the table. This is depicted in the simulation diagram depicted in the figure. \ref{fig:simGrid4x4}.

\begin{table}[htb]
	\caption{State definition for a table partitioned as a grid of size $3 \times 3$. ($0$-unclean grid, $1$- clean grid, $X$ - grid with object)}
	\begin{center}
		\begin{tabular}{|c|c|c|c|c|c|c|c|c|}
			\hline
			\textbf{G9} & \textbf{G8}& \textbf{G7} & \textbf{G6} & \textbf{G5}& \textbf{G4} & \textbf{G3} & \textbf{G2}& \textbf{G1}\\ \hline
			0 & 0 & 0 & 0 & X & 0 & 0 & 0 & 0   \\ \hline
                0 & 0 & 0 & 0 & X & 0 & 0 & 0 & 1  \\ \hline
                0 & 0 & 0 & 0 & X & 0 & 0 & 1 & 0   \\ \hline
                .. & .. & .. & .. & .. & .. & .. & .. & ..  \\ \hline
                .. & .. & .. & .. & .. & .. & .. & .. & ..  \\ \hline
                0 & 0 & 0 & 0 & X & 0 & 1 & 1 & 1  \\ \hline
                0 & 0 & 0 & 0 & X & 1 & 0 & 0 & 0   \\ \hline
                0 & 0 & 0 & 0 & X & 1 & 0 & 0 & 1   \\ \hline
			.. & .. & .. & .. & .. & .. & .. & .. & ..  \\ \hline
                .. & .. & .. & .. & .. & .. & .. & .. & ..  \\ \hline
                1 & 1 & 1 & 1 & X & 1 & 1 & 1 & 0   \\ \hline
                1 & 1 & 1 & 1 & X & 1 & 1 & 1 & 1    \\ \hline
		\end{tabular}	
	\end{center}
	\label{tab:state3x3}
\end{table}

\begin{table*}[htb]
	\caption{State definition for a table partitioned as grids as $4 \times 4$. ($0$-unclean grid, $1$- clean grid, $X$ - grid with object)}
	\begin{center}
		\begin{tabular}{|c|c|c|c|c|c|c|c|c|c|c|c|c|c|c|c|}
			\hline
			\textbf{G16} & \textbf{G15}& \textbf{G14} & \textbf{G13} & \textbf{G12}& \textbf{G11} & \textbf{G10} & \textbf{G9}& \textbf{G8} & \textbf{G7} & \textbf{G6}& \textbf{G5} & \textbf{G4} & \textbf{G3}& \textbf{G2} & \textbf{G1} \\ \hline
			0 & 0 & 0 & 0 & 0 & X & X & 0 & 0 & X & X & 0 & 0 & 0 & 0 & 0  \\ \hline
                0 & 0 & 0 & 0 & 0 & X & X & 0 & 0 & X & X & 0 & 0 & 0 & 0 & 1   \\ \hline
                0 & 0 & 0 & 0 & 0 & X & X & 0 & 0 & X & X & 0 & 0 & 0 & 1 & 0     \\ \hline
                0 & 0 & 0 & 0 & 0 & X & X & 0 & 0 & X & X & 0 & 0 & 0 & 1 & 1 \\ \hline
                .. & .. & .. & .. & .. & .. & .. & .. & ..&.. & .. & .. & .. & .. & .. & ..  \\ \hline
                .. & .. & .. & .. & .. & .. & .. & .. & ..&.. & .. & .. & .. & .. & .. & .. \\ \hline
                .. & .. & .. & .. & .. & .. & .. & .. & ..&.. & .. & .. & .. & .. & .. & .. \\ \hline
			0 & 0 & 0 & 0 & 0 & X & X & 0 & 0 & X & X & 1 & 0 & 0 & 1 & 1   \\ \hline
                0 & 0 & 0 & 0 & 0 & X & X & 0 & 0 & X & X & 1 & 0 & 1 & 0 & 0    \\ \hline
                0 & 0 & 0 & 0 & 0 & X & X & 0 & 0 & X & X & 1 & 0 & 1 & 0 & 1    \\ \hline
                0 & 0 & 0 & 0 & 0 & X & X & 0 & 0 & X & X & 1 & 0 & 1 & 1 & 0  \\ \hline
			.. & .. & .. & .. & .. & .. & .. & .. & ..& .. & .. & .. & .. & .. & .. & .. \\ \hline
                .. & .. & .. & .. & .. & .. & .. & .. & ..&.. & .. & .. & .. & .. & .. & .. \\ \hline
                .. & .. & .. & .. & .. & .. & .. & .. & ..&.. & .. & .. & .. & .. & .. & .. \\ \hline
			1 & 1 & 1 & 1 & 1 & X & X & 1 & 1 & X & X & 1 & 1 & 1 & 0 & 0    \\ \hline
                1 & 1 & 1 & 1 & 1 & X & X & 1 & 1 & X & X & 1 & 1 & 1 & 0 & 1      \\ \hline
                1 & 1 & 1 & 1 & 1 & X & X & 1 & 1 & X & X & 1 & 1 & 1 & 1 & 0       \\ \hline
                1 & 1 & 1 & 1 & 1 & X & X & 1 & 1 & X & X & 1 & 1 & 1 & 1 & 1     \\ \hline
		\end{tabular}	
	\end{center}
	\label{tab:state4x4}
\end{table*}

The state $S$ is a representation of an environment consisting of $G$ grids, which includes objects as well. Hence, each state $S$ is represented as an array of size $G\times1$. Here, the number of possible actions for the agent is equal to the count of the grids. An action is described as $g_n$ is used to direct the agent to clean the $n^{th}$ grid. For $3\times3$ grid the action array is $\{g_1, g_2, g_3 ... g_9\}$ and for $4\times4$ grid the action array is $\{g_1, g_2, g_3 .... g_{16}\}$. Action performed by the agent causes state transition from the current state to the next state. For example the initial state in $3\times3$ is $\{0,0,0,0,X,0,0,0,0\}$. If the action selected is $g_1$ then transition happens to the state mentioned as $\{0,0,0,0,X,0,0,0,1\}$. A state machine is developed to describe all possible state transitions of the $3\times3$ grid scenario is given in table \ref{tab:state3x3}. Here, each state is described by the object position and cleanliness condition of each grid. The object position of the grid is mentioned as $X$. All the other grids can have a value of either $0$ or $1$. A Grid with a value of $0$ means it is unclean. Grid with a value of $1$ means it is clean. In this work, the position of the object is taken to be static. Therefore, the terminal state is the state with all the grid in a clean state except the grid with objects. Similarly initial state of $4\times4$ is $\{0,0,0,0,0,X,X,0,0,X,X,0,0,0,0,0\}$ upon receiving action $g_2$ it transits to the state
$\{0,0,0,0,0,X,$ $X,0,0,X,X,0,0,0,1,0\}$. The possible state transitions for $4\times4$ is given in the table \ref{tab:state4x4}. Action selection is based on the $\epsilon$-greedy method. It chooses an action with the maximum possible reward from the current state. The terminal state is when all the grid on the table is cleaned except those that contain the object. The agent may reach the terminal state in any sequence with respect to the order of the grid chosen to clean the table.

Rewards are given, in such a way that the objective of cleaning all the grids on the table is done. The allotment of reward is given in equation \ref{eq:rewards}. The action of cleaning from the appropriate grid is given a maximum positive reward. A maximum negative reward is given for the agent's attempt to clean a grid with objects. Where a small negative reward is allotted for cleaning an already clean grid to keep the agent from cleaning the same grid.

 \begin{equation} \label{eq:rewards}
  r(t)=\begin{cases}
    +1, & \text{cleaning an unclean grid}.\\
    -1, & \text{cleaning grid with object}. \\
    -0.01, & \text{cleaning grid already clean}
  \end{cases}
  \end{equation}

The table cleaning job is logically implemented using Python programming. The grids are cleaned in a way by one grid at a time. The robotic manipulator with a sponge in its end-effector is mentioned with a grid number. The manipulator cleans that particular grid and receives an appropriate reward as per equation \ref{eq:rewards}.  In this work, the focus is on learning the correct sequence of actions to be performed by the agent to reach the goal. The objective is to reach the goal with maximum success rate and reduced distance moved by the manipulator. Further actuating mechanism for the robotic manipulator for reaching and cleaning a specific grid is taken to be known.

\section{$Q-SD$: $Q$ learning with scaled distance metric algorithm}\label{rlBasic}
Infants learn continuously about different objects in the environment by way of interacting with them. By interaction, different attributes such as colour, shape, and usage are learned. Learning in infants happens using instinct observing the action-effect relations. In the same way, RL learns by repeated interactions with the environment \cite{sutton2018reinforcement}. $Q$-learning is an offline policy algorithm which has better convergence than SARSA \cite{anas2021comparison}. The objective of Q learning, as in any RL is to maximize the reward. $Q$-Learning is a table-based method with states as columns and actions as rows. The main focus is to update the state-action Q-table in each iteration. 
In this work a modified Q learning algorithm namely $Q-SD$ is proposed. $Q-SD$ stands for Q-learning with scaled distance metric. The main advantage of $Q-SD$ algorithm is it maximizes the reward and also helps agent the agent to attain the goal with the minimum distance moved. 

\begin{equation}\label{eq:RL_Q}
\begin{aligned}
Q(s_t,g_{n,t}) <= Q(s_t,g_{n,t}) + \alpha [r_{t+1} +  \gamma  maxQ(s_{t+1},\\g_{n',t+1})  - Q(s_t,g_{n,t})]
\end{aligned}
\end{equation}
\begin{equation}\label{eq:optPolicy}
Q^*(s,a) = \underset \pi{max}  \hspace{0.18cm} Q^\pi(s,a) 
\end{equation}

The $Q$-value update equation for the $Q$-learning algorithm is given by equation \ref{eq:RL_Q}. In equation \ref{eq:RL_Q}, $g_{n,t}$ represents the current action chosen from a set of available actions $G$ that can be performed by the agent on the environment. Similarly, $s_t$ is the current state from a set of states $S$ representing the environment. Among all the states in set $S$, one state $S_T$ is called a terminal state, marking the end of the training episode. In this work, the terminal state is achieved when all the grid without objects is clean. $r_t$ represents the immediate reward received according to a predefined pattern. $\alpha$ is the learning rate, which controls how much the Q-values are updated in each iteration. The discount factor $\gamma$ determines the importance of future rewards in the learning process. With each action performed by the agent, there is a state transition in the environment. Consequently, a positive or negative reward is received as feedback for each action and corresponding state transition. Q-learning has a bootstrapping effect as it relates to two consequent states. SARSA also is having a bootstrapping impact but it waits for the next step to happen for the current Q value update. This makes SARSA algorithm convergence longer compared to Q-learning. The reward is maximized to learn optimum policy as given by the equation \ref{eq:optPolicy}. 

\begin{equation}\label{eq:Q_sd}
\begin{aligned}
Q(s_t,g_{n,t}) <= Q(s_t,g_{n,t}) + \alpha [r_{t+1}  +   \gamma maxQ(s_{t+1},\\ g_{n',t+1}) - Q(s_t,g_{n,t})] - s \times dmetric
\end {aligned}
\end{equation}

\begin{figure}
  \includegraphics[width=\linewidth]{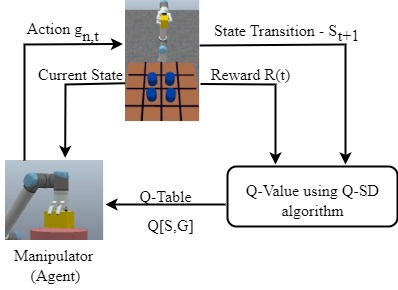}
  \caption{Agent-Environment interface for a $Q-SD$ algorithm}
  \label{fig:RLsuttonBarto}
\end{figure}

In each iteration, the agent is aware of the current state and chooses an action to be executed on the table. Here, the action is the choice grid number $G_n$ where cleaning is to be performed. Action performed causes a state change and consequently, the Q-table $Q[S,A]$ is updated. This updated table is used in the next iteration. The entire flow of sequence is given in Figure. \ref{fig:RLsuttonBarto}.

This work modifies the $Q$-learning equation to incorporate the benefit of minimizing the overall distance moved by the agent. The value update equation for the $Q-SD$ algorithm is given in the equation \ref{eq:Q_sd}. In that equation, a new term $s \times dmetric$ is included in the $Q$-learning equation. Where $dmetric$ is the distance moved by the manipulator from the current grid to the next grid and $'s'$ is the scaling factor constant. The added parameter in the $Q-SD$ algorithm impacts negatively the actual Q-value. Though it reduces the distance moved it tends to pull down the learning of the agent. So the right choice of scaling factor $'s'$ is needed to strike an optimum balance between the agent's task learning and distance moved. Assigning the value of $0$ to $s$ converts the equation of the $Q-SD$ algorithm to a standard $Q$-learning approach. The value of scaling factor $'s'$ varies with the choice and design of the environment in which the agent is placed. The value of scaling factor $'s'$ consideration for $3 \times 3$ grid and $4 \times 4$ grid considered in this work are elaborated in section \ref{simltnResults}.

\begin{algorithm}
\caption{Q-SD Algorithm for task learning with reduced distance movement}\label{alg:Q-SD}
\begin{algorithmic}[1]
 \State \textbf{Input: } $Q[S,G]$, State Table, List of Actions $\{g_1, g_2 .... g_n\}$, Reward $r_{t}$, scaling factor $s \geq 0 $
 \State \textbf{Output: }$Q[S,G]$ table with maximized Q-values
\For{each episode}
        \State Action $g_n \gets \epsilon$-greedyActionSelection$(s_t)$
        \State Perform the action $g_n$ on the table grid
        \State $s_{t+1}  \gets StateTransit(s_t,g_{n,t})$
        \State Reward $r_{t}$ allotted according to equation \ref{eq:rewards}
        \State $dmetric$ = distCalc($g_n, g_{np}$)
        \State Update $Q[S,G]$ using equation \ref{eq:Q_sd}
        \State $itrCnt \gets itrCnt + 1 $ 
        \State $g_{np} \gets g_n $
        \If { $s_{t+1} ==  S_T$ }
            \State End episode
        \ElsIf{ $itrCnt \ge maxItr $ }
            \State End episode
        \ElsIf{ $s_{t+1}$ is a failed State}
            \State $s_{t+1} \gets s_t$
        \EndIf
        \State $s_t \gets s_{t+1}$
\EndFor
\end{algorithmic}
\end{algorithm}

The flow of the $Q-SD$ algorithm is described in algorithm \ref{alg:Q-SD}. Initially, the values of the Q-table $Q[S, G]$ are made as zero. Then, the action is chosen based on the $\epsilon$-greedy method, that is the action with the maximum Q-value for a particular state is chosen. The action performed on the table grid results in state transition. The possible state transitions are described in table \ref{tab:state3x3} and table \ref{tab:state4x4}. The distance calculation between the current grid and the next grid is done using the Euclidean method as elaborated in algorithm \ref{alg:distCalc}. The distance metric and included in the calculation of Q-value. In Algorithm \ref{alg:distCalc}, $G$ represents the total number of grids taken in a scenario. $g_{nc}$ and  $g_{np}$ represent the current and previous grid respectively. 

\begin{algorithm}
\caption{ distCalc - Calculation of distance metric value from a current grid to next grid}\label{alg:distCalc}
\begin{algorithmic}[1]
 \State \textbf{Input: } $g_{nc}$, $g_{np}$
 \State \textbf{Output: } $demtric$  value
 \State $x_c \gets \frac{g_{nc}}{\sqrt{G}} $, $y_c \gets g_{nc} \mod \sqrt{G} $
  \State $x_p \gets \frac{g_{np}}{\sqrt{G}} $, $y_p \gets g_{np} \mod \sqrt{G} $
 \State $ dmetric = \sqrt{(x_c-x_p)^2 + (y_c-y_p)^2}$
\end{algorithmic}
\end{algorithm}

\section{Simulations and Results}\label{simltnResults}
The $Q-SD$ technique was coded in Python and the Coppeliasim software \cite{rohmer2013v} was used for the simulation. In the simulation, the UR10 model was utilised as the manipulator. In this section the results of applying the $Q-SD$ algorithm to an agent cleaning a table which is partitioned as a grid were discussed. Two sets of grid size are taken as input one is $3\times3$ and the other is $4\times4$. Four different graphs were considered to analyze the agent's learning and distance moved. They are the average reward convergence graph, success percentage of the agent, average distance moved with varying weight to distance metric and the graph with normalized values of three different parameters. The first two graphs were used to imply the learning of the agent. The third graph highlights the reduced distance moved by the agent using $Q-SD$ algorithm. The final graph with the normalized value of three different parameters is used to determine the right scale factor for the distance metric to get an optimum balance between success rate and reduced distance moved. 

The agent performing a single action is taken as iteration. The group of iterations is taken as an episode. The episode ends when the terminal state is reached or after a certain amount of iterations.  The collection of episodes is termed as a run, all the readings taken an average of $1000$ such runs. Any table or comparison with a scale factor value of $0$ denotes the use of a standard $Q$-learning algorithm described in equation \ref{eq:RL_Q}.

\subsection{Discussions on impact of $Q-SD$ algorithm on agent's learning and distance moved}
\begin{figure}
 \centering
 \subfigure[$3\times3$ Grid Average Reward]{
    \includegraphics[width=\linewidth]{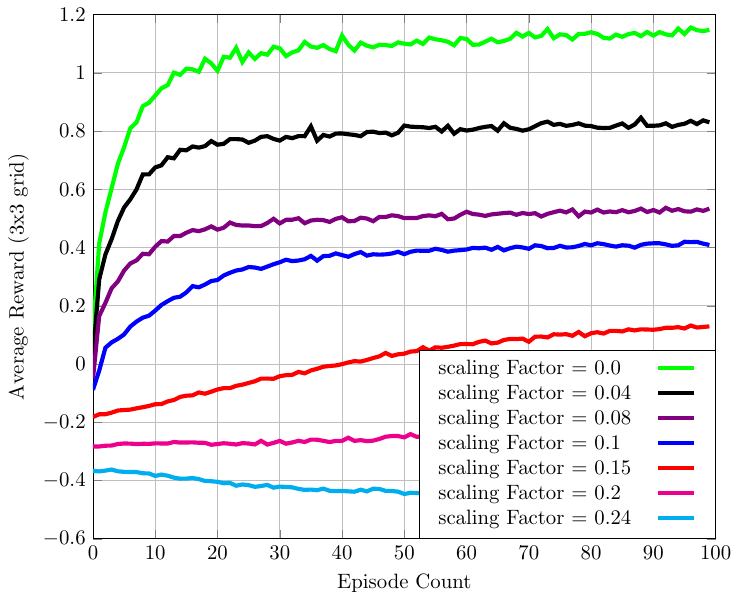}
    \label{fig:avgReward_3x3}
} \\
    \subfigure[$4\times4$ Grid Average Reward]{
    \includegraphics[width=\linewidth]{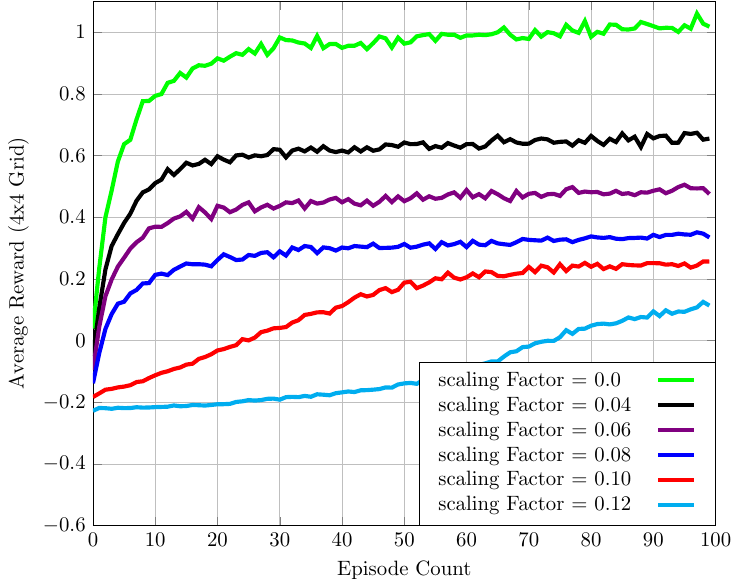}
    \label{fig:avgReward_4x4}
   }
\caption{Convergence of Average Reward }
\label{fig:avgReward}
\end{figure}

In RL the average reward graph illustrates how effectively an agent is learning over the course of time. The average reward graph is depicted in Figure \ref{fig:avgReward}. The average reward converged around 25 episodes in both cases of varying grid size. The convergence of average reward occurs even with variations in the scaling factor to the distance metric. However, it is observed that the increased scale factor to the distance metric impacts negatively the learning rate of the agent. This can be observed in the graph with a negative slope for the average reward with scaling rates of $0.2$ and $0.24$ with a grid size of $3\times3$. In $4\times4$ grid size as well the learning is dragged with scale factors of $0.10$ and $0.12$. Thus, it is observed that the agent's learning is hindered by the enormous magnitude of the scale factor. The high impact of the scaling factor pulls down the $Q$-value resulting in degraded performance.
    
\begin{figure}
 \centering
 \subfigure[Agent's success percentage in $3\times3$ grid with variation in scaling factor of distance metric]{
    \includegraphics[width=\linewidth]{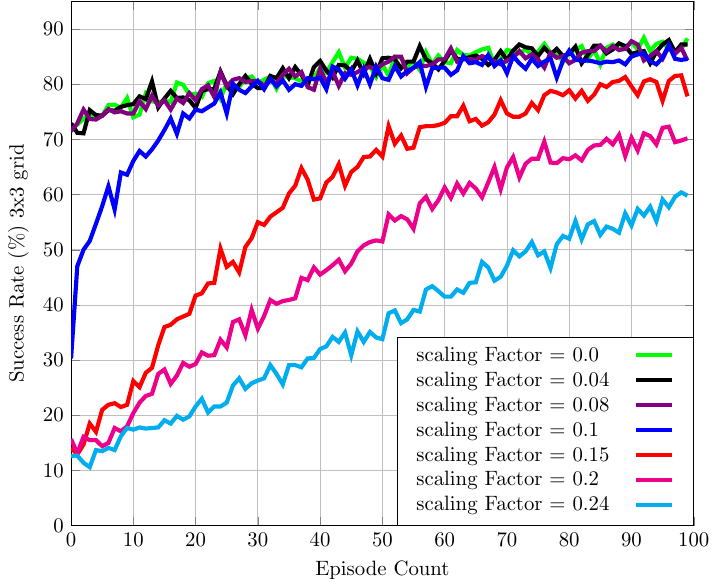}
  \label{fig:success_3x3}
}
    \subfigure[Agent's success percentage in $4\times4$ grid with variation in scaling factor of distance metric]{
    \includegraphics[width=\linewidth]{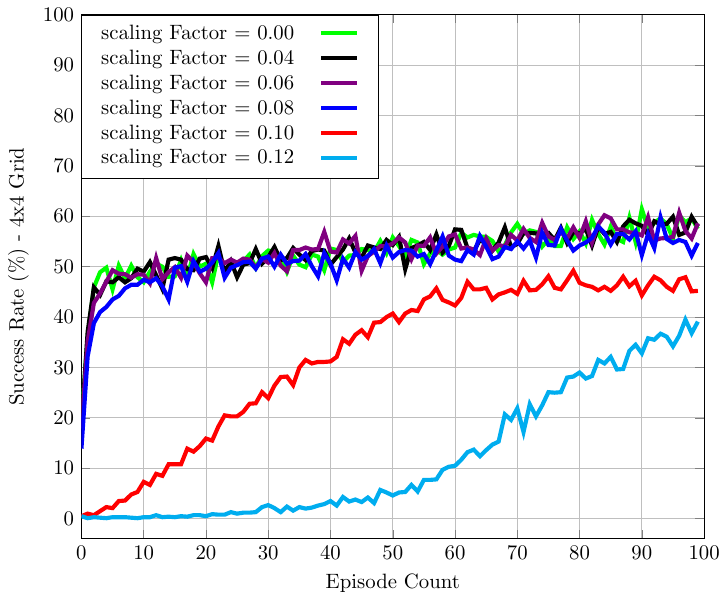}
  \label{fig:success_4x4}
}
\caption{Success Percentage Obtained for different Grid Counts}
\label{fig:success}
\end{figure}

A similar pattern of impact of distance metric and scaling factor on success rate is observed in Figure \ref{fig:success}. As the scaling factor increases it impacts the success rate negatively. Even though the success rate is highest when scaling is not used, the optimum value for the scaling factor is needed to get the highest success rate with the least amount of distance moved by the agent. The maximum success rate obtained for different weight factors to distance metric for $3\times3$ grid size is given in Table \ref{tab:successTable_3x3}. The table mentions the episode count from which the maximum rate of success was reached consistently. As observed with the scale factor of $0.04$ makes the agent learns faster with reduced training episodes. From a scale factor of $0.15$ the distance metric has a negative impact on success rate. Hence higher scale factor forbids the agent in attaining the final goal. The success rate of the agent in a $4 \times 4$ grid with varying scale factor is described in Table \ref{tab:successTable_4x4}. With low scale factor impacts the success rate positively as observed with scale factor value of $0.06$. Here the agent attains maximum success rate with a reduced training episode of $81$. But from a weight factor of $0.10$, the agent's success rate reduces drastically. Moreover, the overall success rate of a $4\times4$ grid is lower compared to a $3\times3$ grid environment due to increased environment complexity.

\begin{table}[htbp]
	\caption{Comparison of Success Rate obtained for different weight to distance metric in a $3\times3$ grid}
	\begin{center}
		\begin{tabular}{|c|c|c|}
			\hline
			\textbf{Scaled} & \textbf{Maximum}& \textbf{Episode} \\ 
                \textbf{Weight} & \textbf{Success Rate (\%)}& \textbf{Count} \\ \hline
			0 & 86 & 81 \\ \hline
                0.04 & 86 & 70 \\ \hline
                0.08 & 86 & 83 \\ \hline
                0.10 & 84 & 79 \\ \hline
                0.15 & 79 & 85 \\ \hline
                0.20 & 68 & 83 \\ \hline
                0.24 & 56 & 89 \\ \hline
		\end{tabular}
	\end{center}
	\label{tab:successTable_3x3}
\end{table}

\begin{figure}
 \centering
 \subfigure[$3\times3$ Average Distance]{
    \includegraphics[width=\linewidth]{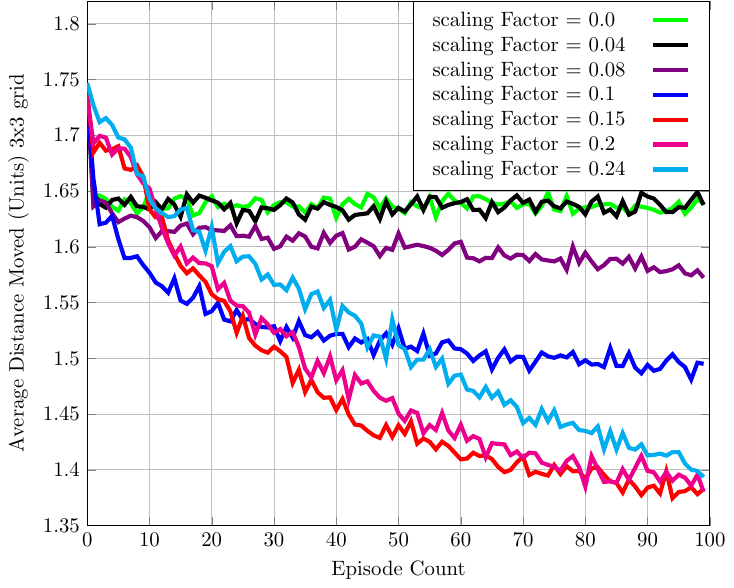}
    \label{fig:avgDist_3x3}
}
\subfigure[$4\times4$ Average Distance]{
   \includegraphics[width=\linewidth]{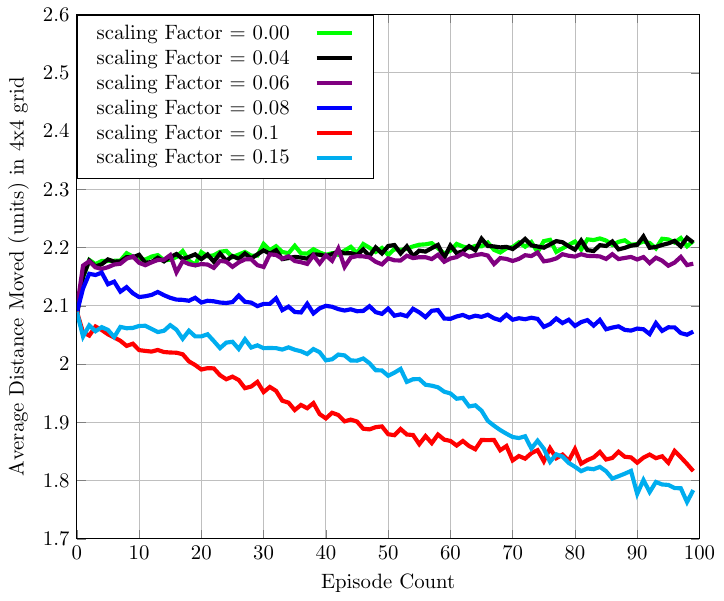}
  \label{fig:avgDist_4x4}
}
\caption{Average distance with change in Weight factor of distance metric}
\label{fig:avgDist}
\end{figure}

\begin{table}[htbp]
	\caption{Comparison of Success Rate obtained for different weight to distance metric in a $4\times4$ grid}
	\begin{center}
		\begin{tabular}{|c|c|c|}
			\hline
			\textbf{Scaled} & \textbf{Maximum}& \textbf{Episode} \\ 
                \textbf{Weight} & \textbf{Success Rate (\%)}& \textbf{Count} \\ \hline
			0 & 58 & 88 \\ \hline
                0.04 & 57 & 88 \\ \hline
                0.06 & 59 & 81 \\ \hline
                0.08 & 56 & 83 \\ \hline
                0.10 & 47 & 89 \\ \hline
                0.12 & 35 & 91 \\ \hline
		\end{tabular}
	\end{center}
	\label{tab:successTable_4x4}
\end{table}
In figure \ref{fig:avgDist} the average distance moved by the agent in each episode is plotted against the episode count. It is computed by dividing the total distance moved by the number of iterations in each episode. It is observed that the $Q-SD$ algorithm has a positive impact in reducing the distance moved by the agent with appropriate value to the scale factor. 

\begin{figure}
 \centering
 \subfigure[Normalized Values in $3\times3$ Grid]{
        \includegraphics[width=\linewidth]{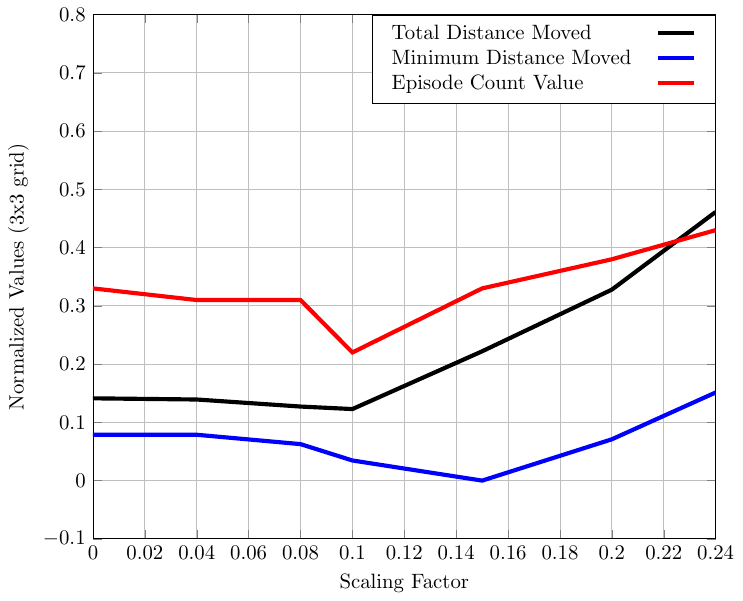}
  \label{fig:normalized_3x3}
}
\subfigure[Normalized Values in $4\times4$ Grid]{
   \includegraphics[width=\linewidth]{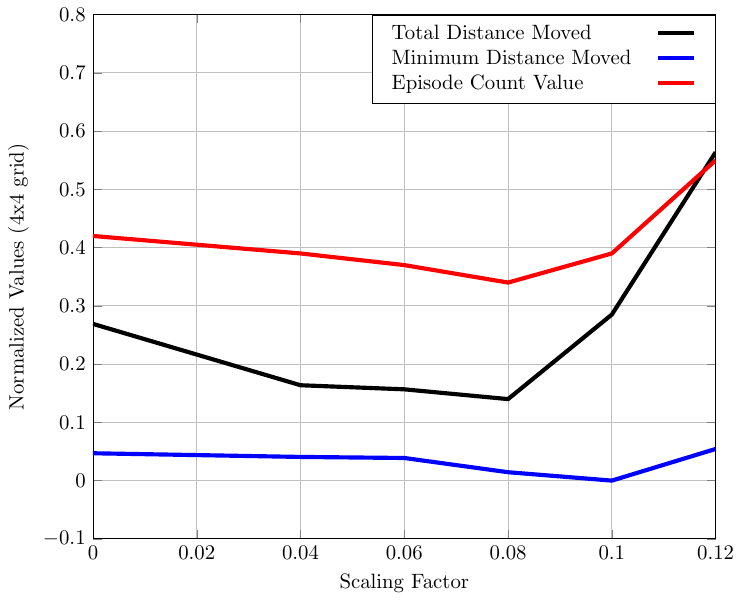}
  \label{fig:normalized_4x4}
}
\caption{Normalized values of Total Distance, Minimum Distance and Episode Count}
\label{fig:normalized}
\end{figure}

The $Q-SD$ algorithm reduces the overall distance moved but impacts the success rate negatively. So, with proper constrained towards the scaling factor of the distance parameter a balance can be struck. The optimum choice of the scale factor is made using three parameters. They are the total distance moved across all episodes, the minimum distance moved in all episodes and the episode count up to which the total distance moved is less than the average distance across all episodes. The values obtained for these values are described in table \ref{tab:scaleIden_3x3} and table \ref{tab:scaleIden_4x4} for $3 \times 3$ and $4 \times 4$ grid dimensions respectively. These values are normalized and plotted in the single graph as shown in Figure \ref{fig:normalized}. The graph has three curves mapped concerning different weight factors. The curve in black represents the total distance moved across all the episodes with respect to the weight factor. The curve in blue represents the minimum value of the total distance moved by the agent in all the episodes. The curve in red represents the count of episodes from which the total distance moved is less than the average total distance over all the episodes. These values are normalized to make them comparable. In Figure \ref{fig:normalized_3x3} dip is observed in the curve of episode count and average of total distance for the weight factor of $0.1$ for a $3 \times 3$ grid size. Hence, for this environment, the weight factor of $0.1$ is the optimal choice to get a reduced distance moved with the least impact on agent success. In the same way in Figure \ref{fig:normalized_4x4}, the dip in the curve is seen for a weight factor of $0.08$. Making it the optimum choice. Similar inference is obtained from Table \ref{tab:successTable_3x3} and Table \ref{tab:successTable_4x4}. There the success rate starts decreasing drastically after a weight factor of $0.1$ and $0.08$ in $3 \times 3$ and $4 \times 4$ grids respectively. Up to those weight factors, there is only a slight variation in success rate is observed.  

\begin{table}[htbp]
	\caption{Appropriate scaling factor identification using total distance, minimum distance and episode count in a $3\times3$ grid}
	\begin{center}
		\begin{tabular}{|c|c|c|c|}
			\hline
			\textbf{Scaled} & \textbf{Total}& \textbf{Minimum} & \textbf{Episode}  \\ 
                \textbf{Weight} & \textbf{Distance}& \textbf{Distance} & \textbf{Count}\\ 
                \textbf{} & \textbf{(units)}& \textbf{(units)} & \textbf{}\\ \hline
			0 & 15.45 & 14.19 & 33\\ \hline
                0.04 & 15.41 & 14.19 & 31 \\ \hline
                0.08 & 15.16 & 13.87 & 31\\ \hline
                0.10 & 15.08 & 13.30 & 22\\ \hline
                0.15 & 17.07 & 12.61 & 33\\ \hline
                0.20 & 19.20 & 14.03 & 38\\ \hline
                0.24 & 21.88 & 15.65 & 43\\ \hline
		\end{tabular}
	\end{center}
	\label{tab:scaleIden_3x3}
\end{table}

\begin{table}[htbp]
	\caption{Appropriate scaling factor identification using total distance, minimum distance and episode count in a $4\times4$ grid}
	\begin{center}
		\begin{tabular}{|c|c|c|c|}
			\hline
			\textbf{Scaled} & \textbf{Total}& \textbf{Minimum} & \textbf{Episode}  \\ 
                \textbf{Weight} & \textbf{Distance}& \textbf{Distance} & \textbf{Count}\\ 
                \textbf{} & \textbf{(units)}& \textbf{(units)} & \textbf{}\\ \hline
			0 & 39.00 & 32.19 & 42\\ \hline
                0.04 & 35.77 & 32.00 & 39 \\ \hline
                0.06 & 35.56 & 31.94 & 37\\ \hline
                0.08 & 35.04 & 31.20 & 34\\ \hline
                0.10 & 39.49 & 30.75 & 39\\ \hline
                0.12 & 48.03 & 32.42 & 55\\ \hline
		\end{tabular}
	\end{center}
	\label{tab:scaleIden_4x4}
\end{table}

As observed in both cases the less significant weight factor boosts the performance of agent's learning. An increase in the scaling factor has a positive impact of reducing the distance moved by the agent. However, if the scaling factor is increased beyond the dipping point it hurts the agent's learning and success. Hence, the optimum choice of weight factor is made to get high success rate with reduced distance moved by the agent. 

\begin{table*}[h]
	\caption{Comparison of different area coverage algorithms}
	\begin{center}
		\begin{tabular}{|p{2cm}|p{3cm}|p{3.5cm}|p{4cm}|}
			\hline
			\textbf{Methodology}& \textbf{Type of Robot} & \textbf{Application}  & \textbf{Utility Value} \\ \hline
                Q-traversal Algorithm \cite{xiao2020distributed}& Multiple Drones & Coverage of physical areas & Covers area with less time. But not reliable in case of distance between drones is large  \\ \hline
                 Deep black RL \cite{lakshmanan2020complete}& Tetromnino robot & Cleaning and maintenance & Minimum energy consumption during morphological transformation \\ \hline 
                 PPO \cite{moon2022path} & Mobile Robot & Cleaning in a mobile robot & Coverage better than conventional zigzag and random methods \\ \hline 
                 Q-learning based CPP \cite{zhang2023predator}& Mobile Robot &  Area coverage by a mobile robot & Covering area with less repetitions \\ \hline
                Non-random targeted viewpoint sampling strategy \cite{glorieux2020coverage}& Robotic Manipulator & Surface metrology by robot inspection  & Decrease in the duration of the lifetime for inspection jobs\\ \hline
                This work & Robotic Manipulator & Cleaning a table (with static objects) & Reduced distance traveled by the manipulator \\ \hline
             
		\end{tabular}	
	\end{center}
	\label{tab:comparison}
\end{table*}
Thus this work proposes a $Q-SD$ algorithm which learns the task with an additional utility value of the reduction in distance moved. This $Q-SD$ algorithm is applied to a task of cleaning a table which is partitioned as grids. Two different grid partitions are considered they are $3\times3$ and $4\times4$. The influence of the distance parameter is controlled by selecting an appropriate scale factor, ensuring that the success rate remains unaffected while reducing the distance moved by the manipulator. A comparison of different algorithms used for area coverage is given in table \ref{tab:comparison}. Each approach is utilised for various applications and possesses a utility value beyond task learning. As in this work, has additional utility value of reduced distance moved in addition to learning the task of cleaning the table. 

\section{Conclusion and Future Work} \label{conclusion}
The current research introduces a novel algorithm called the $Q-SD$ algorithm. The method involves incorporating a scaled distance metric parameter into a typical $Q$-learning algorithm. The $Q-SD$ algorithm is utilised to control a robotic manipulator (agent) to clean a table that is divided into grids. Two configurations of grid partitions were evaluated: one with dimensions of $3\times3$ and another with dimensions of $4\times4$. Furthermore, the arrangement involved the presence of objects occupying a few grids in the central area. In addition to its task learning capabilities, the $Q-SD$ algorithm also gives a utility value. The distance parameter enables the agent to clean the grid systematically, hence minimizing the distance travelled by the agent across several grids. The displacement is decreased by 8.61\% and 6.7\% for $3 \times 3$ and $4 \times 4$ grids, respectively. The weight of the scaling factor is selected appropriately to achieve a balance between the success rate and the distance moved. As observed in section \ref{simltnResults} the increase in grid count resulted in a decrease in success percentage due to environmental complexity. Therefore other algorithms like Deep-Q network can be employed to tackle highly complex environments. Further, mobile manipulators may be employed to achieve a larger coverage area. This work focused on addressing the presence of stationary barriers in the grid. However, there is potential for further enhancement by incorporating dynamic objects and addressing the cleaning of areas located beneath objects. The scaling factor is taken to have a value of $s \geq 0$, and its impact on agent learning for negative values can be investigated further. 


\end{document}